\documentclass[sigconf]{acmart}

\copyrightyear{2026}
\acmYear{2026}
\setcopyright{cc}
\setcctype{by}
\acmConference[CIKM '26]{Proceedings of the 35th ACM International Conference on Information and Knowledge Management}{November 07--11, 2026}{Rome, Italy}
\acmBooktitle{Proceedings of the 35th ACM International Conference on Information and Knowledge Management (CIKM '26), November 07--11, 2026, Rome, Italy}
\acmDOI{10.1145/3799682.3840198}
\acmISBN{979-8-4007-2539-5/2026/11}
\AtBeginDocument{%
  }

\title{SWXAI: A Unified Multimodal Explainability Library for Space Weather Forecasting and Heliophysics}

\author{Benedict Antonio Mervyn}
\affiliation{%
  \institution{Georgia State University}
  \city{Atlanta}
  \state{GA}
  \country{USA}
}
\email{bmervyn1@student.gsu.edu}
\author{Anli Ji}
\affiliation{%
  \institution{California State University, Fullerton}
  \city{Fullerton}
  \state{CA}
  \country{USA}
}
\email{anli.ji@fullerton.edu}
\author{Manolis K. Georgoulis}
\affiliation{%
  \institution{Johns Hopkins University Applied Physics Laboratory}
  \city{Laurel}
  \state{MD}
  \country{USA}
}
\email{manolis.georgoulis@jhuapl.edu}
\author{Berkay Aydin}
\affiliation{%
  \institution{Georgia State University}
  \city{Atlanta}
  \state{GA}
  \country{USA}
}
\email{baydin2@gsu.edu}

\renewcommand{\shortauthors}{Benedict Antonio Mervyn, Anli Ji, Manolis K. Georgoulis, and Berkay Aydin}

\begin{document}

\begin{abstract}
Space weather forecasting and heliophysics increasingly rely on machine learning models that process heterogeneous inputs such as multi-viewpoint multi-wavelength solar images and rasters, multivariate time series, spatial annotations, and tabular event-level data. However, explainability workflows remain fragmented across libraries, model interfaces, and output formats, limiting reuse and reproducibility. We present SWXAI, an open-source library that unifies explainable AI (XAI) workflows for image, time-series, and tabular data in space weather forecasting and heliophysics. SWXAI provides a common, extensible interface across commonly used vision, time-series, and tabular learning models, with validation, adaptation, and persistent explanation objects. It also includes domain-grounded examples, reproducible notebooks, tests, and documentation, enabling users to inspect outputs, swap models or explainers, and extend workflows to new heliophysics datasets. SWXAI reduces modality-specific engineering overhead and supports comparable, reusable explanations across heterogeneous scientific machine learning pipelines. A benchmark against direct backend implementations shows that SWXAI substantially reduces the code users write to produce an explanation, with no practical runtime cost.
\end{abstract}

\keywords{explainable artificial intelligence, heliophysics, space weather forecasting, scientific machine learning}

\ccsdesc[500]{Computing methodologies~Machine learning}
\ccsdesc[300]{Applied computing~Physical sciences and engineering}
\ccsdesc[300]{Information systems~Data mining}

\maketitle
\begin{figure*}[t]
    \centering
    \includegraphics[
        width=\textwidth,
        trim=8 12 8 12,
        clip
    ]{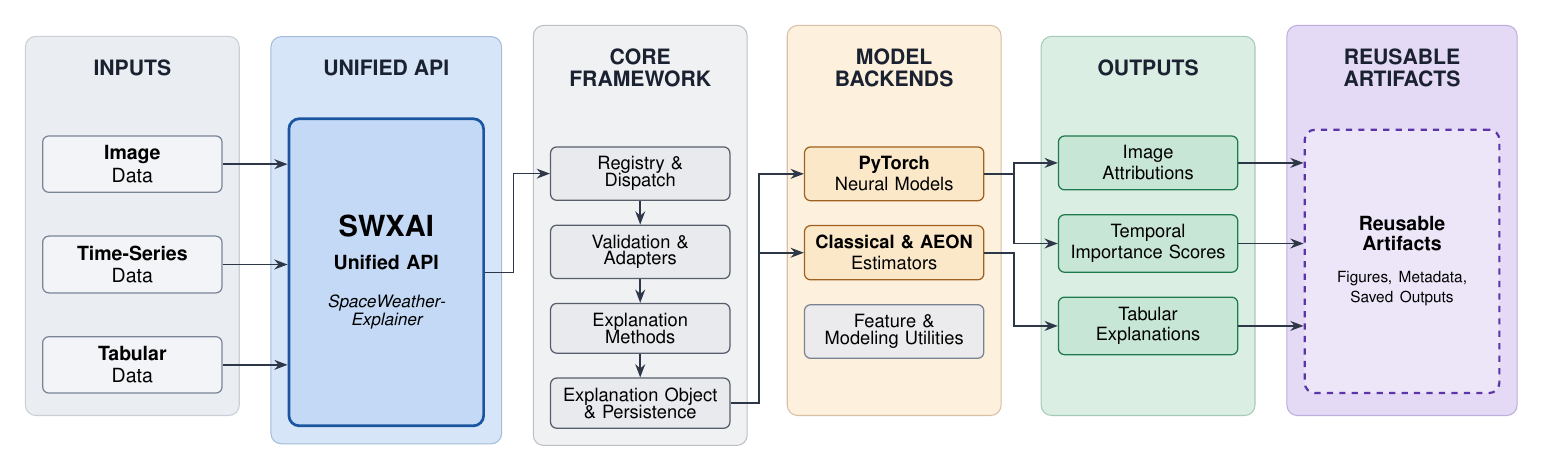}
    \caption{Overview of the SWXAI architecture, showing multimodal inputs, the unified API, registry and validation layers, backend model support, and explanation outputs. Output structure (API schema, visualization templates) is consistent across runs; attribution values are run-specific.}
    \Description{Overview of the SWXAI architecture with multimodal inputs, a unified API, registry and validation layers, model backends, and reusable explanation outputs.}
    \label{fig:architecture}
\end{figure*}

\section{Introduction}

Space weather refers to the dynamic conditions in the near-Earth space environment driven by solar activity, including solar flares, coronal mass ejections (CME), and solar energetic particle (SEP) events, which can disrupt satellite operations, power grids, and communication systems~\cite{Schwenn2006}. Heliophysics is the broader scientific discipline studying the Sun, its influence on the solar system, and the physical processes governing these phenomena~\cite{SS2010hp}. Machine learning is increasingly applied across both domains for tasks such as solar flare prediction, active-region characterization, and image- or time-series-based analyses~\cite{nita2020machine}. As these models become part of scientific workflows, explainability becomes essential. In many settings, it is not enough to traditionally validate models; researchers also need to understand which regions, intervals, or input features drive the model's decisions or predictions.

Despite this need, explanation practices remain fragmented across modalities and models. Existing explainability tools address each modality in isolation, with incompatible interfaces and no shared structure~\cite{BarredoArrieta2020}. This fragmentation increases engineering overhead and limits reusability across projects. Space weather and heliophysics research faces this problem directly, where a single research pipeline may span image, temporal, and structured data modalities~\cite{Han2025}. No existing resource couples explanation methods across image, time-series, and tabular data with the time-series classification backends, domain datasets, and persistent artifacts that heliophysics workflows require.

To address this gap, we present SWXAI, an open-source Python library for unified explainability in space weather forecasting and heliophysics. Rather than targeting a single data modality, SWXAI provides a common interface for explanation generation across image, time-series, and tabular data, supporting gradient-based explainers for deep learning models, model-agnostic explainers for classical estimators, and persistent explanation objects that enable saving, inspection, and cross-pipeline comparison of outputs. The library includes domain-grounded examples spanning active-region patches, multivariate time series, and solar event tabular data, paired with validation utilities, saved explanation artifacts, and tests so users can inspect, reproduce, and adapt results.

The main contributions of this work are: (1) a unified explainability resource spanning image, time-series, and tabular heliophysics workflows with cross-backend support for PyTorch, scikit-learn, and aeon; (2) an integrated framework combining explanation generation, input validation, model adaptation, output persistence, and reusable artifacts; and (3) representative heliophysics datasets, runnable examples, tests, and documentation establishing a practical foundation for reusable interpretability across multimodal space weather applications.

\noindent\textbf{Related Work:} Many general-purpose XAI tools and libraries are available in the literature. Captum provides PyTorch interpretability~\cite{kokhlikyan2020captum}; SHAP~\cite{lundberg2017shap} and LIME~\cite{ribeiro2016lime} support feature-attribution and local explanations; aeon provides unified time-series learning~\cite{middlehurst2024aeon}; and OmniXAI~\cite{omniXAI} offers a unified interface spanning multiple data types and focuses on explanation methods without domain-specific models or datasets. These tools are powerful but require manual pipeline assembly for heliophysics workflows.

XAI has been increasingly adopted within the space weather forecasting community for interpreting predictive models across a range of targets. Post hoc attribution methods, including Guided Grad-CAM, Deep SHAP, and Integrated Gradients, have been applied to flare and CME kinematics prediction models~\cite{yi2021visual,pandey2023ds,sun2022predicting,pandey2023ecml, Gandhi2025}, and sliding window multivariate time series forests (Slim-TSF) have been used for active-region-based flare prediction with feature-level temporal explanations~\cite{ji2023bigdata}. Physics-guided counterfactual explanations have addressed SEP event prediction~\cite{Patil2025}, and white-box explainable AI has extracted human-interpretable conditions for SEP forecasting~\cite{fujita2026sep}, though these efforts remain distributed across incompatible interfaces and output formats.

SWXAI is motivated by the need for a domain-grounded resource that packages explanation methods with the appropriate interfaces, adapters, examples, and reusable outputs needed in heliophysics workflows, connecting these foundational libraries through a common interface for image, time-series, and tabular data, so researchers can move between modalities within the same pipeline without rebuilding modality-specific workflow infrastructure.

\section{SWXAI Framework Overview}

SWXAI is implemented as a Python package with a unified public interface centered on a front-end explainer object. Figure~\ref{fig:architecture} summarizes the framework architecture. A user specifies an explanation method, modality, and task, and the framework dispatches the request to the appropriate backend implementation. This design hides much of the modality-specific and backend-specific complexity from the user and keeps the framework extensible.

Method-specific code is isolated from the public API so new model frameworks can be added without changing user-facing calls. The framework consists of five main components. The Unified API Layer exposes \texttt{Space\-Weather\-Explainer}, which routes each request by explanation method, modality, and task type. The Registry and Dispatch Layer organizes methods by name and decouples the public API from individual implementations. The Validation and Adaptation Layer normalizes inputs and bridges PyTorch models~\cite{paszke2019pytorch} with scikit-learn estimators and aeon classifiers~\cite{pedregosa2011sklearn,middlehurst2024aeon}. The Explanation Object and Persistence Layer packages attribution values, run metadata, and figures into a consistent output structure that can be saved and compared across runs. Finally, Feature and Modeling Utilities transform raw temporal observations into explainable feature representations for downstream heliophysics models. Shape or backend mismatches raise descriptive errors.

\section{Resource Contents and Supported Workflows}

\begin{figure}[b]
    \centering
    \includegraphics[width=1.00\columnwidth,height=5.75cm,keepaspectratio,trim=20.7 20.7 20.7 20.7,clip]{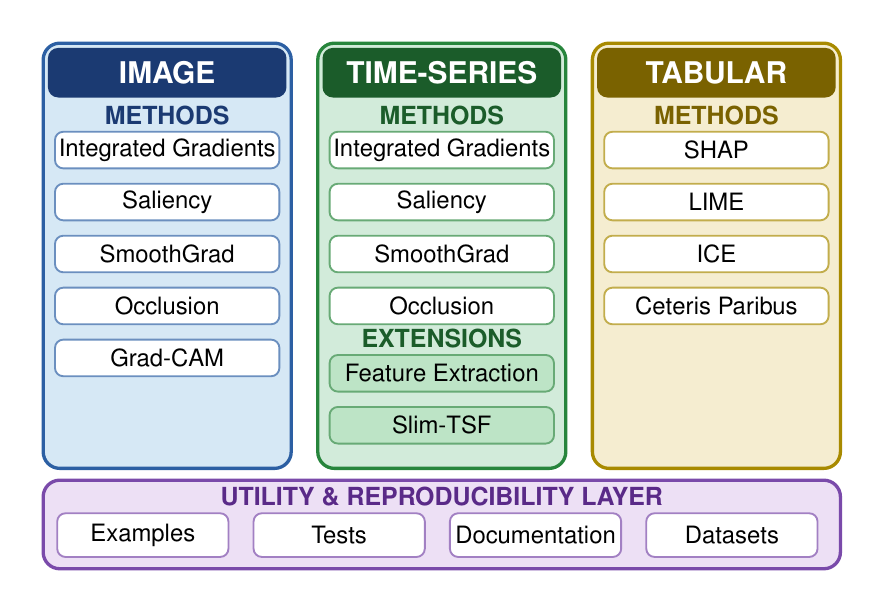}
    \caption{SWXAI coverage across data modalities, explanation methods, modeling and feature-extraction extensions, and utility and reproducibility resources.}
    \Description{Coverage figure for SWXAI showing supported image, time-series, and tabular explanation methods, plus extensions and reproducibility resources.}
    \label{fig:capability}
\end{figure}

SWXAI supports explainability across the main data types encountered in heliophysics workflows while preserving a consistent interface across model backends. We present an illustration of currently supported modalities and explanations in Figure~\ref{fig:capability}. For image-based workflows, SWXAI supports gradient-based and perturbation-based explainers for PyTorch~\cite{paszke2019pytorch} models, including Integrated Gradients~\cite{sundararajan2017axiomatic}, Saliency~\cite{simonyan2014deep}, SmoothGrad~\cite{smilkov2017smoothgrad}, Occlusion, and Grad-CAM-style methods~\cite{selvaraju2017gradcam}. These methods make the framework suitable for spatial attribution over 2D input, such as magnetograms~\cite{hoeksema2014hmi} or EUV image products~\cite{lemen2012aia}.

For time-series workflows, SWXAI supports both gradient-based explainers for PyTorch~\cite{paszke2019pytorch} time-series models and model-agnostic occlusion-based explanations for scikit-learn estimators and aeon classifiers~\cite{pedregosa2011sklearn,middlehurst2024aeon}. SWXAI also includes time-series feature extraction utilities and Slim-TSF, a custom interval-based modeling component for feature-based space weather forecasting workflows~\cite{ji2024slimtsf}; Slim-TSF supports swappable base estimators including Random Forest~\cite{breiman2001rf}, Extra Trees, and XGBoost~\cite{chen2016xgboost}, with a hybrid embedded feature selection approach~\cite{ji2024hybrid}, and the framework additionally bundles a configurable transformer-based time-series classifier for attention-based temporal attribution. Five feature extractors (MiniRocket~\cite{dempster2021minirocket}, DWT, SFA-Fast~\cite{schafer2012sfa}, TSFresh~\cite{christ2018tsfresh}, and RSAST~\cite{rojas2023rsast}) are provided to convert raw temporal sequences into tabular feature representations compatible with scikit-learn estimators~\cite{pedregosa2011sklearn}.

For tabular workflows, SWXAI supports SHAP-based explanations~\cite{lundberg2017shap}, LIME-based local explanations~\cite{ribeiro2016lime}, Individual Conditional Expectation (ICE)~\cite{goldstein2015peeking}, and Ceteris Paribus profiles~\cite{kuzba2019ceteris}.

The repository bundles three datasets, one per modality. The HMI active-region patch data~\cite{pandey2024hmiar} provides 2{,}000 line-of-sight magnetogram patches from 864 unique HARP regions: 334 flaring (X- and M-class) and 1{,}666 non-flaring (A-, B-, and C-class). An undersampled SWAN-SF subset~\cite{angryk2020swan} provides 300 multivariate windows across five partitions (30 flaring, 30 non-flaring each), each with 60 timesteps of 55 magnetic-field parameters at 12-minute cadence over 167 active regions. The geomagnetic storms dataset provides 132 event-level records from the NASA CDAW Dst$\leq\!-100$~nT catalog~\cite{cdaw2024geostorms} spanning 1996--2025, comprising 116 coronal mass ejection (CME)-, 11 stream interaction region (SIR)-, and 5 compound CME-SIR-driven events, of which 40 reach severe intensity at Dst$\leq\!-150$~nT. Each dataset illustrates a modality-specific explanation workflow; full provenance and loading instructions are documented in the repository.

\section{Implementation, Availability, and Reproducibility}

SWXAI hides modality-specific and backend-specific complexity behind a consistent interface. For PyTorch-based image and time-series workflows, it converts inputs into tensors, validates expected shapes, and checks model-input compatibility before calling the selected explanation method. For non-PyTorch estimators, it exposes prediction-function interfaces for black-box explainers, so users can switch model classes without rewriting the surrounding pipeline.

The standardization of the explanation object is a deliberate design choice to improve reproducibility. Every explainer in SWXAI returns the same container structure: an \texttt{Explanation} dataclass with three fields: \texttt{attributions} (the raw attribution array), \texttt{summary} (a dictionary capturing method settings, modality/task information where applicable, and attribution statistics), and \texttt{figures} (a dictionary of generated plots), so users can compare outputs across methods and runs, log artifacts to persistent storage, and inspect saved explanations without modifying the surrounding pipeline. Importantly, what contributes to reusability is the workflow structure (API calls, output schema, visualization templates), which remains consistent across runs, models, and datasets; the actual attribution values and plot contents are run-specific. The registry pattern supports extensibility: each explainer is implemented independently and registered by name, so new methods can be added without touching the public interface.

\noindent\textbf{Availability:} 
The source repository is available at \url{https://bitbucket.org/gsudmlab/swxai} under the GNU GPL-3.0 license, and documentation is available at \url{https://swxai.readthedocs.io}.
The repository contains the package, datasets, modality-specific end-to-end notebooks, documentation, validation benchmark, and tests for reproducing and extending the workflows described in this paper; installation instructions are provided in the README.

\noindent \textbf{Resource Validation:} We benchmarked three bundled workflows (HMI Grad-CAM, SWAN-SF occlusion, geomagnetic-storm SHAP, detailed in Sec.~\ref{sec:outputs}) against direct implementations on the underlying backends, each producing the same artifacts (i.e., attribution array, run metadata, and figure). Counting executable statements at the explanation stage from the abstract syntax tree, we find that the direct workflows require 12, 21, and 23 user-authored statements against 2 for SWXAI in every case (83--91\% fewer); manual integration operations fall from 8, 11, and 8 to 2, and persistence calls from 3, 2, and 3 to 1. For the time-series workflow, no existing library provides model-agnostic occlusion over scikit-learn and aeon classifiers, so the direct baseline requires reimplementing the method itself. Attributions match the direct implementations exactly, so the interface introduces no numerical modification, and saved artifacts reload identically. Across 30 paired runs with warm-up and bootstrap confidence intervals, runtime differences remain below two milliseconds in all three workflows. We note that these workflows are demonstrations of library capability rather than full-scale training pipelines; benchmark scripts, raw timing logs, and environment metadata are in the repository.

\section{Representative Outputs and Discussion} \label{sec:outputs}
\begin{figure}[t!]
    \centering
    \includegraphics[width=1.0\columnwidth,height=4.75cm]{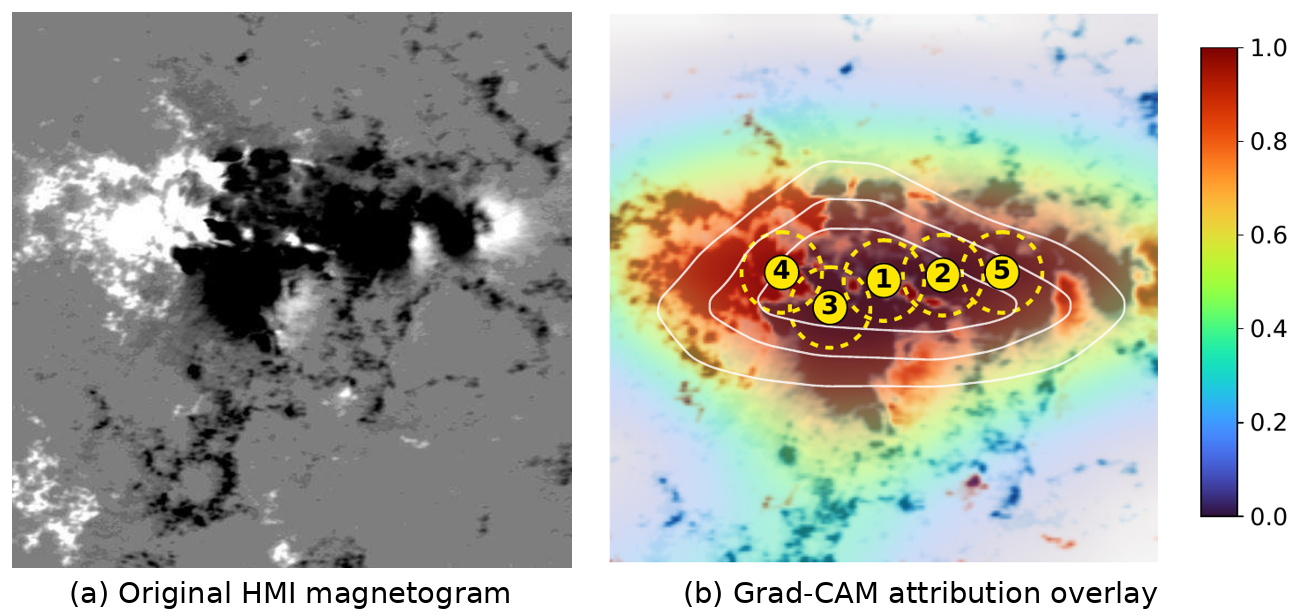}
    \caption{Original HMI active-region magnetogram and Grad-CAM explanation with top-5 high-importance locations for HARP~\#856 (Label: M1.6).}
    \Description{Two-panel figure showing the original HMI active-region magnetogram on the left and the Grad-CAM attribution overlay with a colorbar and top-5 attribution points on the right for HARP 856.}
    \label{fig:image_example}
\end{figure}

\begin{figure}[t!]
    \centering
    \includegraphics[width=0.98\columnwidth]{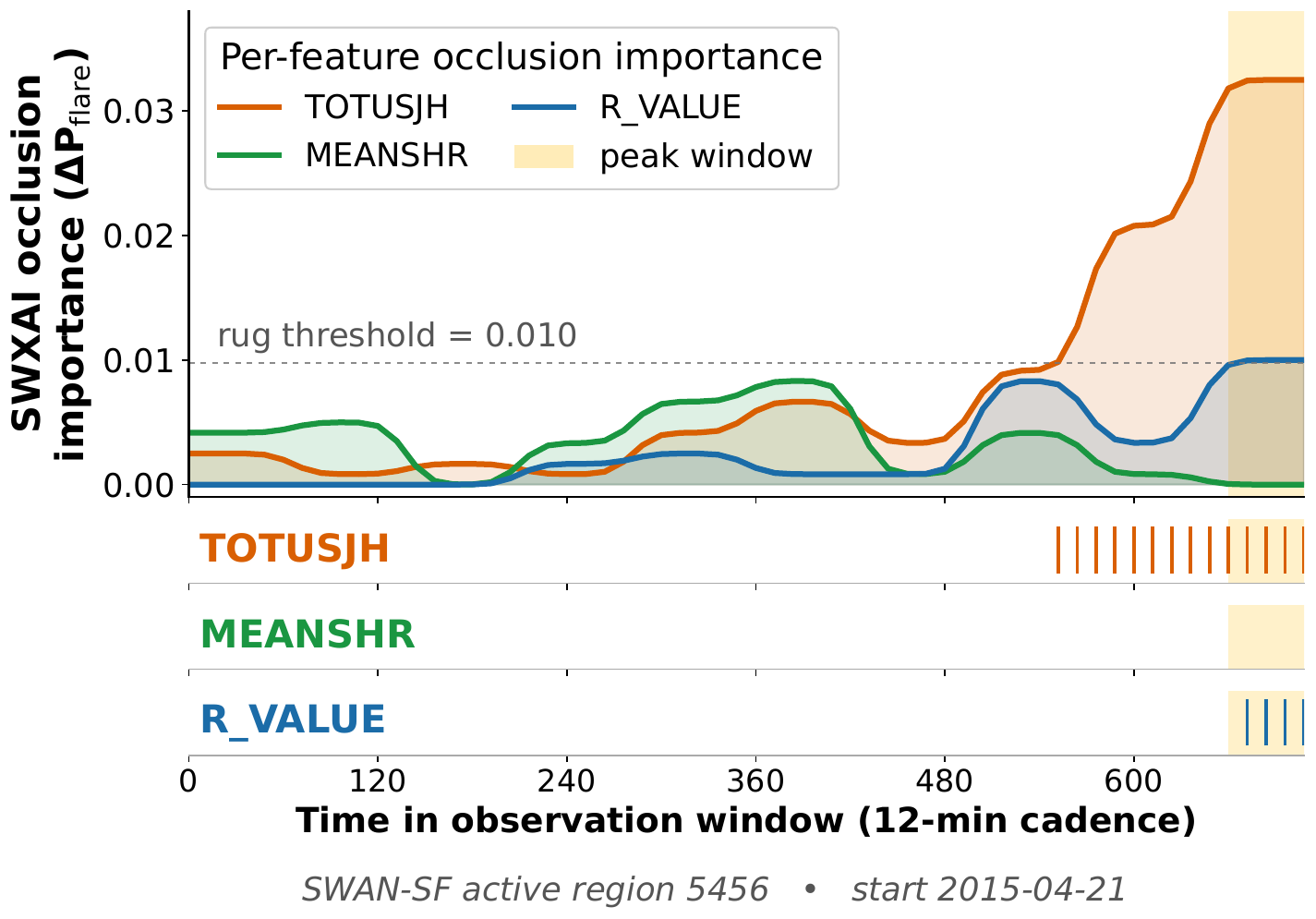}
    \caption{SWXAI occlusion importance for a SWAN-SF active-region instance (HARPNUM~5456). Per-feature $\Delta P_{\text{flare}}$ curves are shown for TOTUSJH, MEANSHR, and R\_VALUE across the observation window. Light per-curve shading emphasizes the magnitude under each feature curve, the vertical shaded band marks the highest aggregate-importance peak window, and rug strips identify time steps above the attribution threshold.}
    \Description{Temporal occlusion importance visualization for HARP 5456, showing per-feature importance curves, a shaded peak window, and rug strip markers.}
    \label{fig:timeseries_example}
\end{figure}

\begin{figure}[t!]
    \centering
    \includegraphics[width=1.00\columnwidth,height=5.02cm]{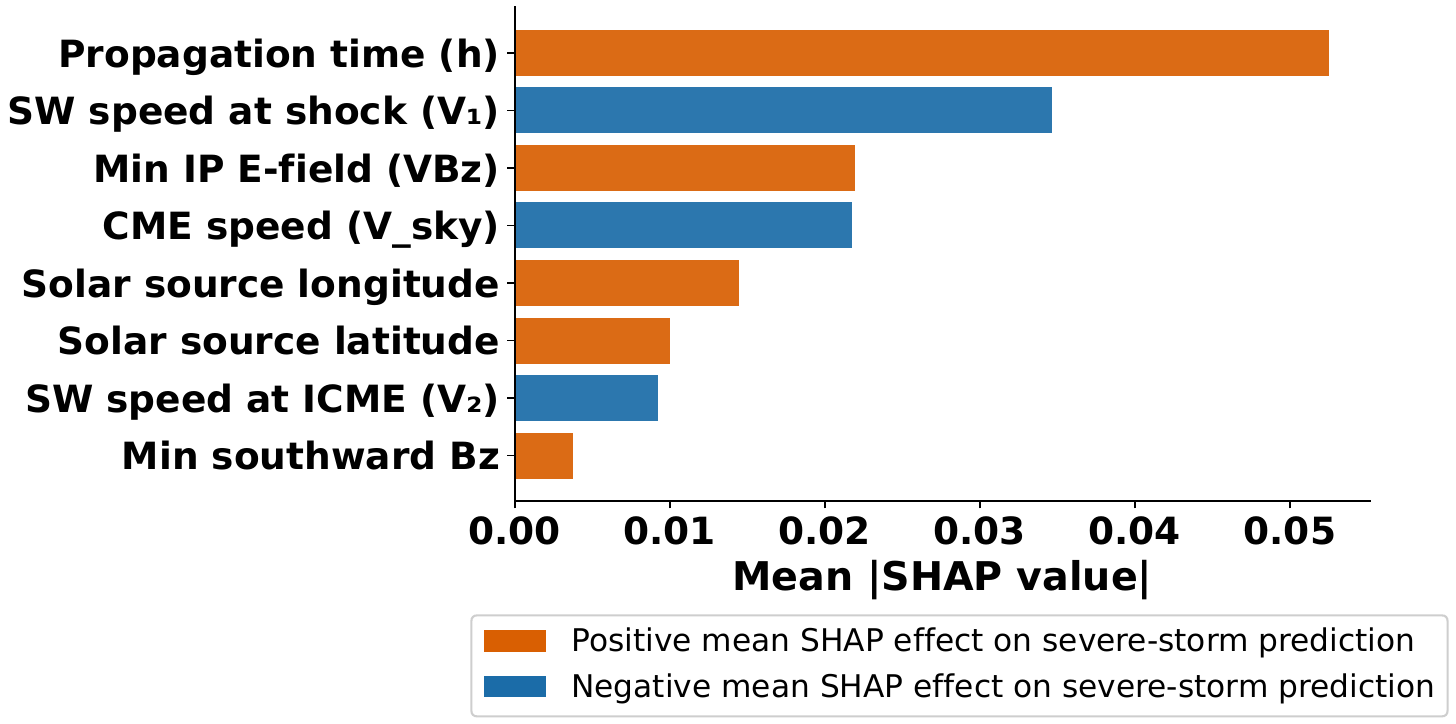}
    \caption{SHAP feature attribution on the geomagnetic storms dataset, ranked by mean absolute SHAP value. The prediction target is severe-storm occurrence (Dst\,$\leq$\,$-150$\,nT); orange bars denote features that increase the predicted storm probability (positive mean SHAP effect) and blue bars denote features that decrease it (negative mean SHAP effect).}
    \Description{SHAP-based feature attribution on the geomagnetic storms dataset showing ranked feature importances with positive and negative mean SHAP effects.}
    \label{fig:tabular_example}
\end{figure}

Figures~\ref{fig:image_example}--\ref{fig:tabular_example} present representative SWXAI outputs across the three modalities, each following the same process: (1) preparation of modality-specific data, (2) loading a relatively lightweight model, (3) invoking \texttt{SpaceWeather\-Explainer}, and (4) exporting outputs for reuse.

For the image modality, Figure~\ref{fig:image_example} shows a Grad-CAM explanation for a flaring instance in the HMI active-region patch dataset using a PyTorch flare prediction model, pairing the original magnetogram with the attribution overlay and the ranked locations of concentrated spatial contribution. Other inputs, models, or attribution methods can be substituted while preserving the same explanation call, validation path, and artifact-saving behavior.

For the time-series modality, we train an interval-based classifier on SWAN-SF multivariate series and explain it via occlusion. The novel, customizable visualization shows which channels, and at which intervals, most influence the flare prediction, capturing the cross-channel temporal reasoning central to space weather forecasting.

For the tabular modality, geomagnetic storm event data demonstrates event-level feature attributions (SHAP in Figure~\ref{fig:tabular_example}) across solar-wind, CME, and source-location variables, identifying which properties most influence the severity classification. As with the other modalities, the target definition, feature set, model, and explainer configuration are stored together for reuse with a different dataset, classifier, or method.

The bundled notebooks reproduce these modality-specific examples end to end and retain the resulting attributions, metadata, and figures as reusable artifacts.

Beyond these representative outputs, SWXAI contributes by introducing interval-based feature construction, five configurable feature extractors, and model-agnostic explanation coverage across scikit-learn and aeon classifiers. The architecture itself is domain-independent; the datasets, models, and visual conventions are not. This extends coverage to classical and interval-based models, beyond the PyTorch-only scope of libraries such as Captum. The current release prioritizes system integration over new algorithms, with some explainers requiring backend-specific capabilities handled through validation and adapters.

\section{Conclusion and Future Work}
We presented SWXAI, a unified multimodal explainability resource for space weather forecasting and heliophysics, providing a consistent interface across image, time-series, and tabular workflows while supporting both deep-learning and classical estimators. SWXAI reduces engineering overhead and supports reproducible interpretability workflows for heterogeneous scientific data. Future work will expand SWXAI with additional explanation methods~\cite{Patil2025, Hong2025}, benchmarking utilities~\cite{Adeyeha2024}, broader heliophysics datasets~\cite{Khani2026, Rotti2022}, newer multimodal architectures, and validation of attributions with heliophysics domain experts.

\begin{acks}
This work was supported by the National Science Foundation under Grants No. 2513886 and 2513887. The authors thank the public heliophysics data providers whose datasets support the examples in this work.
\end{acks}

\section*{GenAI Usage Disclosure}
Generative AI tools were used only for grammar, formatting, and minor code polishing. All technical content and contributions were developed, reviewed, and verified by the authors.

\bibliographystyle{ACM-Reference-Format}
\bibliography{refs}

\end{document}